# Dual Representation Learning for One-Step Clustering of Multi-View Data

Wei Zhang, Zhaohong Deng, *Senior Member*, *IEEE*, Kup-Sze Choi, Jun Wang, *Member*, *IEEE*, Shitong Wang

***Abstract*—Multi-view data are commonly encountered in data mining applications. Effective extraction of information from multi-view data requires specific design of clustering methods to cater for data with multiple views, which is non-trivial and challenging. In this paper, we propose a novel one-step multi-view clustering method by exploiting the dual representation of both the common and specific information of different views. The motivation originates from the rationale that multi-view data contain not only the consistent knowledge between views but also the unique knowledge of each view. Meanwhile, to make the representation learning more specific to the clustering task, a one-step learning framework is proposed to integrate representation learning and clustering partition as a whole. With this framework, the representation learning and clustering partition mutually benefit each other, which effectively improves the clustering performance. Results from extensive experiments conducted on benchmark multi-view datasets clearly demonstrate the superiority of the proposed method.**

## I. Introduction

Advance of data acquisition technology enables the collection of data from multiple sources and with multiple representations. This kind of data is called multi-view data. For example, a news story can be written in different languages and each language version can be regarded as a view.

To make full use of multi-view data, multi-view clustering, as an important part of multi-view learning, has received great attention in recent years. The two main categories of multi-view clustering methods are the *original views-based methods* and the *common latent view-based methods*.

Original views-based methods: This type of methods usually extends the conventional clustering algorithms, like K-Means, fuzzy clustering, or spectral clustering, to the corresponding multi-view versions [1-6]. A typical framework of this type of methods is shown in Fig. 1(a) and the representative methods are discussed below. By introducing variable weights, a two-level variable weighting multi-view clustering method called TW-k-Means was proposed [1], in which the weights were assigned to both features and views to represent their importance in clustering procedure. Based on kernel Canonical Correlation Analysis (CCA), Houthuys et al. proposed a multi-view kernel spectral clustering method [2]. Cleuziou introduced collaborative learning to the fuzzy membership matrix of each view and proposed a multi-view collaborative fuzzy clustering method Co-FKM [3]. Combining the entropy weighting mechanism and collaborative learning, Jiang et al. proposed a

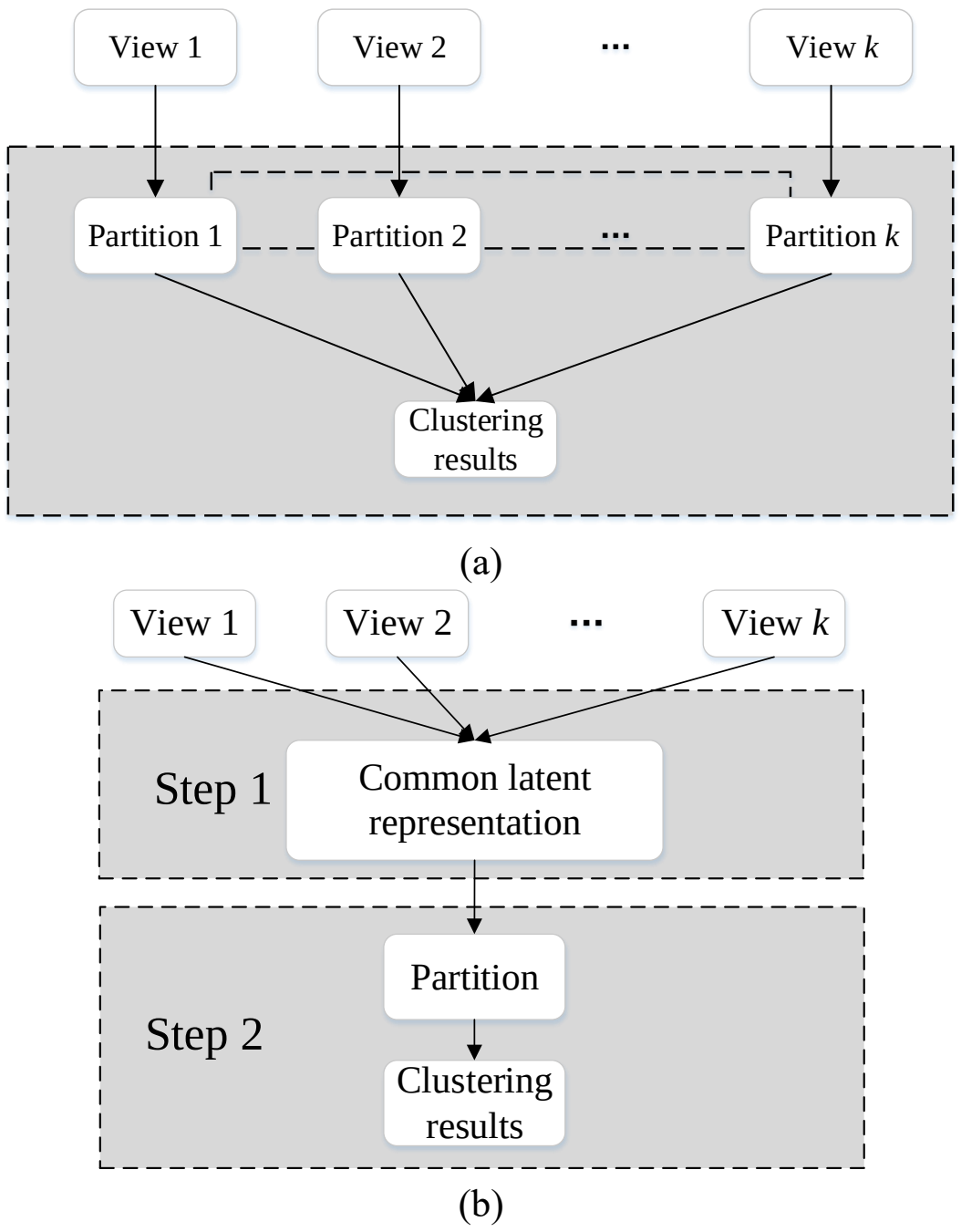

Fig. 1: Typical frameworks of two main multi-view clustering methods: (a) original views-based multi-view clustering methods; (b) common latent view-based methods.

This work was supported in part by the NSFC (61772239, 62176105), the Six Talent Peaks Project in Jiangsu Province (XYDXX-056), the Hong Kong Research Grants Council (PolyU 152006/19E), the Hong Kong Innovation and Technology Fund (MRP/015/18) and the Shanghai Municipal Science and Technology Major Project (2018SHZDZX01). (Corresponding author: Zhaohong Deng).

W. Zhang, S. T. Wang are with the School of Artificial Intelligence and Computer Science, Jiangnan University and Jiangsu Key Laboratory of Media Design and Software Technology, Wuxi 214122, China (e-mail: 7201607004@stu.jiangnan.edu.cn; wxwangst@aliyun.com).

Z. Deng is with the School of Artificial Intelligence and Computer Science, Jiangnan University, Wuxi 214122, China, and Key Laboratory of Computational Neuroscience and Brain-Inspired Intelligence (LCNBI) and ZJLab, Shanghai 200433, China. (e-mail: dengzhaohong@jiangnan.edu.cn).

K. S. Choi is with The Centre for Smart Health, the Hong Kong Polytechnic University, Hong Kong (e-mail: thomasks.choi@polyu.edu.hk).

J. Wang is with the Key laboratory of Specialty Fiber Optics and Optical Access Networks, Joint International Research Laboratory of Specialty Fiber Optics and Advanced Communication, Shanghai Institute for Advanced Communication and Data Science, School of Communication and Information Engineering, Shanghai University, Shanghai, 200444, China (e-mail: wangjun_shu@shu.edu.cn).

weighted view collaborative fuzzy c-means method (WV-Co-FCM) [4]. Based on the exemplar invariant assumption, Zhang et al. proposed a multi-view and multi-exemplar fuzzy clustering method, which divided a multi-view dataset into groups with multiple exemplars [5]. Based on co-training technique, Kumar et al. proposed a multi-view spectral clustering method by propagating complementary information from one view to others [6].

Common latent view-based methods: This type of methods usually uses representation learning techniques, such as self-representation, matrix factorization and canonical correlational analysis, to learn a common latent view of the original views for clustering [7-13]. A typical framework of this type of multi-view clustering methods is shown in Fig. 1(b) and the representative methods are discussed below. Liu et al. proposed a method using non-negative matrix factorization (NMF) to reconstruct a common representation of all original views, and then apply K-Means algorithms to the common representation to get the final clustering results [7]. Zhang et al. employed self-representation learning to reconstruct the common representation for spectral clustering [8]. Yin et al. proposed a multi-view clustering method which achieved a sparse representation of each high-dimensional data point with respect to other data points in the same view. It also maximized the correlation between the representations of different views [9]. Based on self-representation learning, Cao et al. explored the complementary information with the Hilbert-Schmidt independence criterion and proposed a new multi-view clustering method [10]. To address data nonlinearity, Andrew et al. proposed a deep CCA method to explore the common representation for multi-view clustering [11]. Wang et al. proposed a deep multi-view subspace clustering algorithm by combining the global and local structures with the self-expression layer [12]. By transforming the original features in the Euclidean space to the low-dimensional binary Hamming space, a new binary multi-view clustering method was proposed [13].

Although great advance in the two types of multi-view clustering methods has been achieved, there are still critical challenges to overcome. The two key challenges are:

(1) In multi-view data, there exist not only the consistent knowledge shared by different views, but also a set of unique knowledge belonging to the different views. In the two types of methods motioned above, the first type focuses on exploring the unique knowledge of each view and the second type focuses on exploring the consistent knowledge. Simultaneous exploration and utilization of both types is still a challenge.

(2) For the second type of methods, i.e., Common latent view-based methods, most of them separate representation learning from clustering partition, which precludes the interaction of the two processes. The resulted representation learning is therefore not pertinent to a given clustering task. Hence, effective one-step learning mechanisms that integrate two processes deserve in-depth investigation.

Attempts have been made in recent years to tackle the first challenge mentioned above. For example, Luo et al. explored the common and specific information simultaneously using self-representation and proposed a multi-view subspace clustering method [14]. Zhou et al. projected the original views into low-dimensional representation and then used self-representation learning to explore the common and specific information from the low-dimensional representation for clustering [15]. To fully utilize the common and specific information of multi-view data, Deng et al. used the NMF to learn the common representation, and then combined the common hidden view with the original views to realize a visible-hidden view collaboration multi-view clustering method [16]. Although these methods [14-16] have alleviated the challenge to some extent, they still suffer from several obvious weaknesses. First, all these methods separate clustering partition from representation learning, and therefore the representation learned lacks the pertinence for clustering. Second, most of these methods [14], [15] only combine the learned common and specific part as a whole for clustering, which cannot take advantage of the two parts. Third, most of these methods do not consider the redundancy between the types of two knowledge in the clustering partition, which degrades the clustering performance. Finally, most of these methods are based on self-representation learning, which are not large-dataset friendly [17]. The above analyses indicate that a new efficient method is needed to explore the consistent and unique knowledge for multi-view clustering.

Efforts have also been made to meet the second challenge discussed above. One-step methods for multi-view clustering have received attention in recent years. For example, Wang et al. proposed a one-step method integrating representation learning with partition clustering guided by the pseudo label [17]. Zhang et al. utilized self-representation learning to learn the common representation of multi-view data and integrated it with spectral rotation to directly achieve the clustering results [18]. Similarly, Tang et al. combined self-representation learning and partition clustering into one process, and further introduced low-rank tensor constraints to improve the discriminability of the common view [19]. Although these methods [17-19] have alleviated the second challenge, there is one main issue needs to be solved: when one-step learning is implemented, these methods focus on exploring the consistent information but ignore other information in the multi-view data, such as the specific information of each view. Besides, the difference between each of the views is ignored [18], [19] and the complex initialization is needed [17]. So, there is still a large room for improvement for the existing one-step multi-view clustering methods.

Based on the above analyses, it is clear that an efficient one-step multi-view clustering method with the ability to explore the consistent and unique knowledge is in demand. To this end, we propose a One-step Multi-view Clustering method by exploring Dual Representation (OMC-DR) in this paper. First, a new multi-view dual representation learning mechanism is proposed based on a novel combination of matrix factorizations [20], [21], which can learn the common representation of different views and the specific representation of each view simultaneously. Benefitting from the efficiency of matrix factorization, the proposed dual representation learning is more efficient for large dataset than the commonly used self-representation-based mechanism in other methods [14], [15]. In particular, an orthogonal constraint is designed to reduce the redundant information between the common and specific representations. Then, to make the representation learning more pertinent for clustering, a unified optimization framework integrating clustering partition with dual representation is proposed to realize one-step clustering. In this framework, the cluster indicator matrix can be directly acquired based on the learned common and specific representations, while the discriminability of the learned common and specific representations can also be improved using the information of the clustering partition. Therefore, these two processes mutually benefit each other and lead to more promising clustering performance. Finally, the mechanism of maximum entropy is introduced to balance the influence of different views. By this mechanism, the importance of common and specific representations can be adaptively adjusted in the clustering process.

The main contributions of this paper are summarized as follows:

1) A new dual representation learning mechanism that can simultaneously explore the consistent knowledge of different views and the unique knowledge of each view is proposed.

2) A new one-step multi-view clustering method that unifies common and specific representation learning, and cluster indicator matrix learning into an adaptive framework is proposed.

3) Extensive experiments on real-world datasets are carried out to validate the high efficiency and effectiveness of the proposed method.

The remainder is of this paper organized as follows. The related work is briefly described in Section II. In Section III, the proposed method is described in detail. Experimental studies on various datasets are reported in Section IV. Finally, conclusion and future work are given in Section V.

## II. Related Work

Common representation learning is the key of multi-view clustering methods based on common latent view. For multi-view clustering methods based on the original views, K means clustering is usually used as the base model to develop the multi-view versions. In this study, common representation learning and K-Means clustering are adopted as the basis of the proposed method. We briefly introduce them in this section.

### A. Common Representation Learning

Given a multi-view data $\{\mathbf{X}^k \in R^{N\times d^k}, k = 1,2,\dots,K\}$, where $N$ is the number of samples, $K$ is the number of views, and $d^k$ is the number of features in the $k$-th view. Based on the assumption that the multi-view data have a latent common representation, some methods have been proposed to learn the common representation in recent years, like CCA based methods [22], [23], self-representation based methods [8], [24] and matrix factorization based methods [25]-[28]. Among them, matrix factorization based methods are related to our work and are briefly reviewed as follows. The methods learn the common representation by optimizing the objective function:

$$\min_{\mathbf{H},\mathbf{W}^k} \sum_{k=1}^{K} \|\mathbf{X}^k - \mathbf{H}\mathbf{W}^k\|_F^2 \tag{1}$$

where $\mathbf{H} \in R^{N\times m_c}$ is the common latent representation of all original views, $m_c$ is the number of features in the common latent space, and $\mathbf{W}^k \in R^{m_c\times d^k}$ is the mapping matrix of the $k$-th view. By building a strong connection between the views, the common representation $\mathbf{H}$ can capture the consistent information of different views. By alternating optimization, the update rules of $\mathbf{W}^k$ and $\mathbf{H}$ can be obtained as follows:

$$\mathbf{H} = \sum_{k=1}^{K} \left(\mathbf{X}^k {\mathbf{W}^k}^{\mathrm{T}}\right)\left(\mathbf{W}^k {\mathbf{W}^k}^{\mathrm{T}}\right)^{-1} \tag{2}$$

$$\mathbf{W}^k = (\mathbf{H}^{\mathrm{T}}\mathbf{H})^{-1}(\mathbf{H}^{\mathrm{T}}\mathbf{X}^k) \tag{3}$$

The final common representation is obtained by iteratively implementing the above update rules. Then, we can get the clustering results by feeding $\mathbf{H}$ into K-Means or other traditional clustering algorithms. Obviously, this strategy only explores the consistent knowledge of different views and ignores the unique knowledge of each view. Moreover, for a given clustering task, the learned common representation may not be strongly related to the clustering task.

### B. K-Means Clustering

K-Means is a classical clustering algorithm that is simple and efficient [29-31]. Given a dataset $\mathbf{X} \in R^{N\times d}$, the goal of the K-Means is to find a partition matrix $\mathbf{U} \in R^{C\times N}$ and a cluster center matrix $\mathbf{V} \in R^{d\times C}$, where $C$ is the number of clusters, $N$ is the number of samples and $d$ is the number of features. The objective function of the classical K-Means is defined as follows:

$$\min_{\mathbf{U},\mathbf{V}} \sum_{i=1}^{c}\sum_{j=1}^{N} \mathbf{U}_{i,j} \left\|\mathbf{X}_{j,:} - \mathbf{V}_{:,i}^{\mathrm{T}}\right\|^2$$
$$s.t. \mathbf{U}_{i,j} \in \{0,1\}, \sum_{i=1}^{C} \mathbf{U}_{i,j} = 1 \tag{4}$$

where $\mathbf{V}_{:,i}$ is the center of the $i$-th cluster, $\mathbf{U}_{i,j}$ is the membership of the $j$-th sample to the $i$-th cluster, $\mathbf{X}_{j,:}$ is the $j$-th instance. The equivalence between K-Means clustering and matrix factorization has been shown in [32], i.e., (4) can be transformed as follows:

$$\min_{\mathbf{U},\mathbf{V}} \|\mathbf{X}^{\mathrm{T}} - \mathbf{V}\mathbf{U}\|_F^2 \quad s.t. \mathbf{U}_{i,j} \in \{0,1\}, \sum_{i=1}^{C} \mathbf{U}_{i,j} = 1 \tag{5}$$

Using alternating optimization, we can obtain the update rules of $\mathbf{U}$ and $\mathbf{V}$ as follows:

$$\mathbf{V} = (\mathbf{U}\mathbf{X})^{\mathrm{T}}(\mathbf{U}\mathbf{U}^{\mathrm{T}})^{-1} \tag{6}$$

$$\mathbf{U}_{i,j} = \begin{cases} 1, & i = \underset{c}{\operatorname{argmin}} \left\|\mathbf{X}_{j,:}^{\mathrm{T}} - \mathbf{V}_{:,c}\right\|_F^2 \\ 0, & otherwise \end{cases} \tag{7}$$

## III. MULTI-VIEW DUAL REPRESENTATION LEARNING FOR CLUSTERING

In this section, a new one-step multi-view clustering method based on dual representation learning is proposed to address the two critical challenges discussed in the Introduction. The framework of our method is shown in Fig. 2. This method not only explores the consistent knowledge of different views, but also the unique knowledge of each view. Moreover, the dual representation learning and cluster indicator matrix learning are implemented in a one-step manner to enable the two processes to benefit from each other.

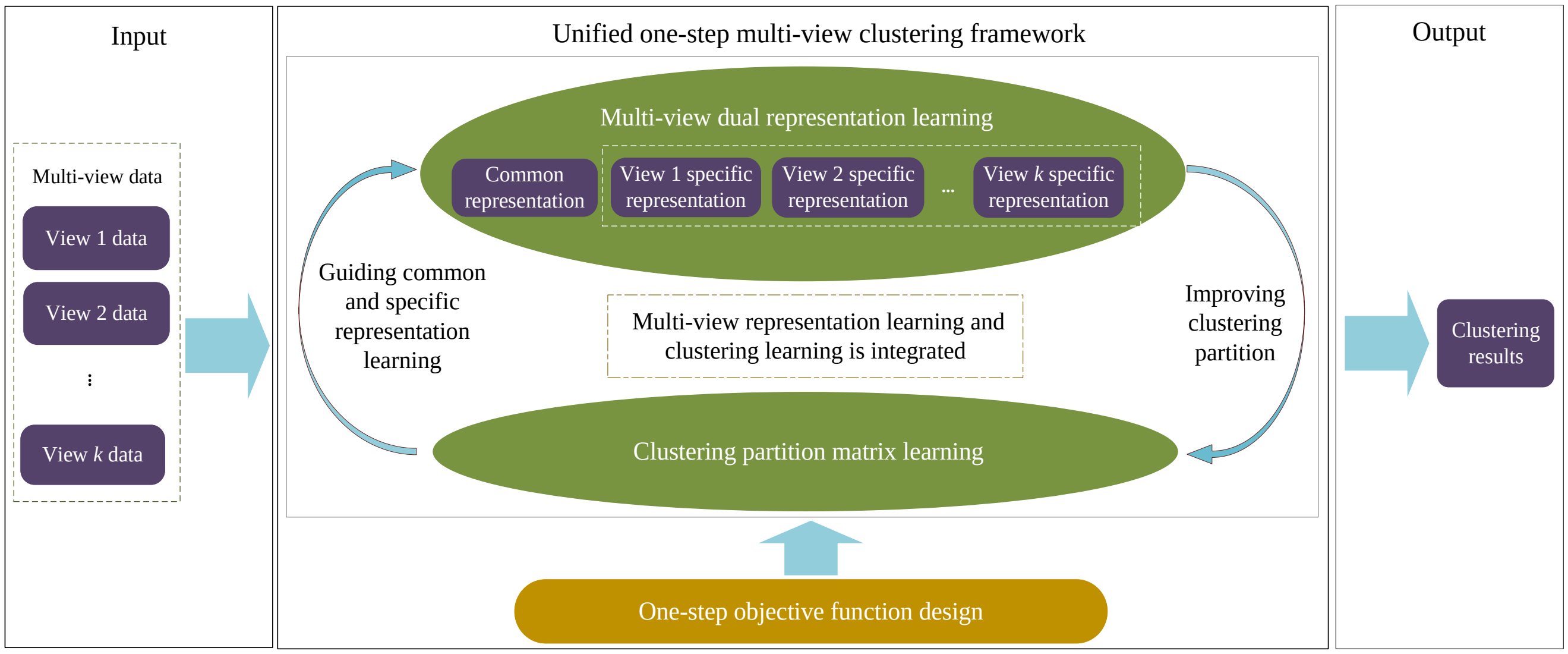

Fig. 2. The framework of the proposed one-step multi-view clustering method integrating dual representation learning and clustering.

### *A. Dual Representation Learning Mechanism for Multi-view Data*

All representation learning based methods assume that there is a common representation, i.e., the consistent knowledge, of different views. However, multi-view data not only contain the consistent knowledge of all the original views, but also the specific discriminant knowledge of each view. Since these two types of knowledge are both useful for clustering partition, traditional common latent representation-based multi-view clustering methods are insufficient to mine multi-view data. Although some self-representation learning based methods have been proposed to explore the consistent and unique knowledge for multi-view clustering [14], [15], they are not large dataset friendly and do not consider the redundancy between the two types of knowledge. Therefore, a more efficient mechanism is needed to explore the consistent and unique knowledge.

Given a multi-view dataset $\{\mathbf{X}^k \in R^{N\times d^k}, k = 1,2,\dots,K\}$ as defined in Section II-A, we propose a dual representation learning mechanism to explore the common and specific representations simultaneously as follows:

$$\min_{\mathbf{H},\mathbf{W}^k,\mathbf{S}^k,\mathbf{P}^k} \sum_{k=1}^{K}\|\mathbf{X}^k - \mathbf{H}\mathbf{W}^k - \mathbf{S}^k\mathbf{P}^k\|_F^2 + \gamma\|\mathbf{H}\|_{2,1}^2 + \gamma\sum_{k=1}^{K}\|\mathbf{S}^k\|_{2,1}^2 + \beta\sum_{k=1}^{K}\|\mathbf{H}^{\mathrm{T}}\mathbf{S}^k\|_F^2 \quad (8)$$

where $\mathbf{H}$ is the common latent representation of different views, $\mathbf{S}^k \in R^{N\times m_s}$ is the specific representation of the $k$-th view, $m_s$ is the number of features in the specific representation. For simplicity, the dimensions of specific representations of different views are set to be same. $\mathbf{P}^k \in R^{m_s\times d^k}$ is the mapping matrix of $k$-th view, $\gamma, \beta > 0$ are two regularization parameters and $\|*\|_F$ is the Frobenius norm.

In (8), the common latent representation of different views and the specific representation of each view are learned simultaneously. To enable the learned representations to be more robust against noise and outliers, the $\|*\|_{2,1}$ norm is introduced for $\mathbf{H}$ and $\{\mathbf{S}^k\}$ [14]. Further, we note that since the common and specific representations are learned simultaneously, the redundant information may exist between them. Inspired by [33], an orthogonal constraint, i.e., $\|\mathbf{H}^{\mathrm{T}}\mathbf{S}^k\|_F^2$, is designed to reduce the redundancy.

### *B. A Novel Objective Function for One-step Multi-view Clustering*

A direct way to perform clustering using the learned common and specific representations is to combine them as a whole with traditional clustering algorithms, such as K-means or spectral clustering [14]. This is a convenient approach but cannot use the discriminability information in the common and specific representations sufficiently. Besides, it separates representation learning from clustering partition without allowing the two processes to benefit from each other. Although there have been several one-step methods designed to realize representation learning and clustering partition simultaneously, these methods only focus on the common representation and ignore the specific representation.

Based on the above analyses, we propose a new one-step learning framework to integrate representation learning and clustering partition, along with a maximum entropy mechanism to adjust the importance of each view. For convenience, we denote the common representation as the $K$+1 view. Finally, the proposed objective function for one-step multi-view clustering is formulated as follows:

$$\min_{\Upsilon} \sum_{k=1}^{K}\|\mathbf{X}^k - \mathbf{H}\mathbf{W}^k - \mathbf{S}^k\mathbf{P}^k\|_F^2 + \gamma\big(\|\mathbf{H}\|_{2,1}^2 + \sum_{k=1}^{K}\|\mathbf{S}^k\|_{2,1}^2\big) + \beta\sum_{k=1}^{K}\|\mathbf{H}^{\mathrm{T}}\mathbf{S}^k\|_F^2 + \sum_{k=1}^{K}\alpha^k\|\mathbf{S}^{k,\mathrm{T}} - \mathbf{V}^k\mathbf{U}\|_F^2 + \alpha^{K+1}\|\mathbf{H}^{\mathrm{T}} - \mathbf{V}^{K+1}\mathbf{U}\|_F^2 + \delta\sum_{k=1}^{K+1}\alpha^k\ln\alpha^k$$
$$s.t.\ \alpha^k \geq 0, \sum_{k=1}^{K+1}\alpha^k = 1, \mathbf{U}_{i,j} \in \{0,1\}, \sum_{i=1}^{C}\mathbf{U}_{i,j} = 1 \quad (9)$$

where $\mathbf{V}^{K+1} \in R^{m_c\times C}$ and $\mathbf{V}^k \in R^{m_s\times C}$ ($k = 1,2,\dots K$) are the clustering center in the space of common and specific representations, respectively. $\mathbf{U} \in R^{C\times N}$ is the cluster indicator matrix shared by $K$ specific views (i.e., $\mathbf{S}^k, k =$ $1,2,\dots K$) and one common view (i.e., $\mathbf{H}$). If the $j$-th sample belongs to the $i$-th class, $\mathbf{U}_{i,j}=1$, otherwise, $\mathbf{U}_{i,j}=0$. $\alpha^k$ is the view weight of the $k$-th view. $\beta$, $\gamma$ and $\delta$ are used to balance the influence of different terms. $\Upsilon=\{\mathbf{H},\mathbf{W}^k,\mathbf{S}^k,\mathbf{P}^k,\mathbf{U},\mathbf{V}^k,\alpha^k\}$ is the set of parameters to be optimized. Further details about (9) are explained below:

1) The first term $\sum_{k=1}^{K}\|\mathbf{X}^k-\mathbf{H}\mathbf{W}^k-\mathbf{S}^k\mathbf{P}^k\|_F^2$ is used to learn the common representation of different views and specific representation of each view.

2) The second term $\|\mathbf{H}\|_{2,1}^2+\sum_{k=1}^{K}\|\mathbf{S}^k\|_{2,1}^2$ is used to ensure robustness of the learned common and specific representations against noise and outliers.

3) The third term $\sum_{k=1}^{K}\|\mathbf{H}^{\mathrm{T}}\mathbf{S}^k\|_F^2$ is used to reduce the redundancy between the common and specific representations.

4) The fourth and fifth terms $\sum_{k=1}^{K}\alpha^k\|\mathbf{S}^{k,\mathrm{T}}-\mathbf{V}^k\mathbf{U}\|_F^2+\alpha^{K+1}\|\mathbf{H}^{\mathrm{T}}-\mathbf{V}^{K+1}\mathbf{U}\|_F^2$ are used to cluster the common representation $\mathbf{H}$ and the specific representation $\mathbf{S}^k$.

5) The sixth term $\sum_{k=1}^{K+1}\alpha^k\ln\alpha^k$ is the entropy of the weights of different views, which is used to adjust the importance of the views adaptively.

In (9), the learned common and specific representations can improve the clustering performance, and in turn, the learned cluster indicator matrix can improve the discriminability of the common and specific representations. This is the mutual self-taught mechanism in the proposed one-step framework.

## C. Optimization

In this subsection, we introduce an alternating optimization scheme to solve the problem in (9).

**Step 1: Update $\mathbf{H}$.** Fixing all variables except $\mathbf{H}$, we obtain the following minimization problem:

$$\min_{\mathbf{H}}\sum_{k=1}^{K}\|\mathbf{X}^k-\mathbf{H}\mathbf{W}^k-\mathbf{S}^k\mathbf{P}^k\|_F^2+\gamma\|\mathbf{H}\|_{2,1}^2+\beta\sum_{k=1}^{K}\|\mathbf{H}^{\mathrm{T}}\mathbf{S}^k\|_F^2+\alpha^{K+1}\|\mathbf{H}^{\mathrm{T}}-\mathbf{V}^{K+1}\mathbf{U}\|_F^2 \quad (10)$$

Taking the derivative of (10) with respect to $\mathbf{H}$ and setting it to zero, the update rule for $\mathbf{H}$ is given by:

$$\mathbf{H}=\left((1+\alpha^{K+1})\mathbf{I}+\beta\sum_{k}^{K}\mathbf{S}^k\mathbf{S}^{k,\mathrm{T}}+\gamma\mathbf{D}\right)^{-1}\left(\sum_{k=1}^{K}(\mathbf{X}^k\mathbf{W}^{k,\mathrm{T}}-\mathbf{S}^k\mathbf{P}^k\mathbf{W}^{k,\mathrm{T}})+\alpha^{K+1}\mathbf{U}^{\mathrm{T}}\mathbf{V}^{K+1,\mathrm{T}}\right) \quad (11)$$

where $\mathbf{D}$ is a diagonal matrix with the $i$-th diagonal value given by $\mathbf{D}_{i,i}=1/\|\mathbf{H}_{i,:}\|_2$.

**Step 2: Update $\mathbf{S}^k$.** Fixing all variables except $\mathbf{S}^k$, we obtain the following minimization problem:

$$\min_{\mathbf{S}^k}\sum_{k=1}^{K}\|\mathbf{X}^k-\mathbf{H}\mathbf{W}^k-\mathbf{S}^k\mathbf{P}^k\|_F^2+\gamma\sum_{k=1}^{K}\|\mathbf{S}^k\|_{2,1}^2+\beta\sum_{k=1}^{K}\|\mathbf{H}^{\mathrm{T}}\mathbf{S}^k\|_F^2+\sum_{k=1}^{K}\alpha^k\|\mathbf{S}^{k,\mathrm{T}}-\mathbf{V}^k\mathbf{U}\|_F^2 \quad (12)$$

Taking the derivative of (12) with respect to $\mathbf{S}^k$ and setting it to zero, the update rule for $\mathbf{S}^k$ is given by:

$$\mathbf{S}^k=\left((1+\alpha^k)\mathbf{I}+\beta\mathbf{H}\mathbf{H}^{\mathrm{T}}+\gamma\mathbf{E}^k\right)^{-1}(\mathbf{X}^k\mathbf{P}^{k,\mathrm{T}}-\mathbf{H}\mathbf{W}^k\mathbf{P}^{k,\mathrm{T}}+\alpha^k\mathbf{U}^{\mathrm{T}}\mathbf{V}^{k,\mathrm{T}}),\ k=1,2,\dots K \quad (13)$$

where $\mathbf{E}^k$ is a diagonal matrix with the $i$-th diagonal value given by $\mathbf{E}_{i,i}^k=1/\|\mathbf{S}_{i,:}^k\|_2$.

**Step 3: Update $\mathbf{W}^k$.** Fixing all variables except $\mathbf{W}^k$, we obtain the following minimization problem:

$$\min_{\mathbf{W}^k}\sum_{k=1}^{K}\|\mathbf{X}^k-\mathbf{H}\mathbf{W}^k-\mathbf{S}^k\mathbf{P}^k\|_F^2 \quad (14)$$

Taking the derivative of (14) with respect to $\mathbf{W}^k$ and setting it to zero, the update rule for $\mathbf{W}^k$ is given by:

$$\mathbf{W}^k=(\mathbf{H}^{\mathrm{T}}\mathbf{H})^{-1}(\mathbf{H}^{\mathrm{T}}\mathbf{X}^k-\mathbf{H}^{\mathrm{T}}\mathbf{S}^k\mathbf{P}^k),\ k=1,2,\dots K \quad (15)$$

**Step 4: Update $\mathbf{P}^k$.** Fixing all variables except $\mathbf{P}^k$, we obtain the following minimization problem:

$$\min_{\mathbf{P}^k}\sum_{k=1}^{K}\|\mathbf{X}^k-\mathbf{H}\mathbf{W}^k-\mathbf{S}^k\mathbf{P}^k\|_F^2 \quad (16)$$

Taking the derivative of (16) with respect to $\mathbf{P}^k$ and setting it to zero, the update rule for $\mathbf{P}^k$ is given by:

$$\mathbf{P}^k=(\mathbf{S}^{k,\mathrm{T}}\mathbf{S}^k)^{-1}(\mathbf{S}^{k,\mathrm{T}}\mathbf{X}^k-\mathbf{S}^{k,\mathrm{T}}\mathbf{H}\mathbf{W}^k),\ k=1,2,\dots K \quad (17)$$

**Step 5: Update $\alpha^k$.** Fixing all variables except $\alpha^k$, we obtain the following minimization problem:

$$\min_{\alpha^k}\sum_{k=1}^{K}\alpha^k\|\mathbf{S}^{k,\mathrm{T}}-\mathbf{V}^k\mathbf{U}\|_F^2+\alpha^{K+1}\|\mathbf{H}^{\mathrm{T}}-\mathbf{V}^{K+1}\mathbf{U}\|_F^2+\delta\sum_{k=1}^{K+1}\alpha^k\ln\alpha^k$$
$$s.t.\ 0<\alpha^k<1,\sum_{k=1}^{K+1}\alpha^k=1 \quad (18)$$

Denoting $\mathbf{S}^{K+1}=\mathbf{H}$, taking the derivative of (18) with respect to $\alpha^k$ and setting it to zero, the update rule for $\alpha^k$ is given by:

$$\alpha^k=\frac{exp\left(-\|\mathbf{S}^{k,\mathrm{T}}-\mathbf{V}^k\mathbf{U}\|_F^2/\delta\right)}{\sum_{l}^{K+1}exp\left(-\|\mathbf{S}^{l,\mathrm{T}}-\mathbf{V}^l\mathbf{U}\|_F^2/\delta\right)},\ k=1,2,\dots K+1 \quad (19)$$

**Step 6: Update $\mathbf{V}^k$.** Fixing all variables except $\mathbf{V}^k$, we obtain the following minimization problem:

$$\min_{\mathbf{V}^k}\sum_{k=1}^{K}\alpha^k\|\mathbf{S}^{k,\mathrm{T}}-\mathbf{V}^k\mathbf{U}\|_F^2+\alpha^{K+1}\|\mathbf{H}^{\mathrm{T}}-\mathbf{V}^{K+1}\mathbf{U}\|_F^2 \quad (20)$$

Denoting $\mathbf{S}^{K+1}=\mathbf{H}$, taking the derivative of (20) with respect to $\mathbf{V}^k$ and setting it to zero, the update rule for $\mathbf{V}^k$ is given by:

$$\mathbf{V}^k=(\mathbf{U}\mathbf{S}^k)^{\mathrm{T}}(\mathbf{U}\mathbf{U}^{\mathrm{T}})^{-1},\ k=1,2,\dots K+1 \quad (21)$$

**Step 7: Update $\mathbf{U}$.** Fixing all variables except $\mathbf{U}$, we obtain:

$$\min_{\mathbf{U}}\sum_{k=1}^{K}\alpha^k\|\mathbf{S}^{k,\mathrm{T}}-\mathbf{V}^k\mathbf{U}\|_F^2+\alpha^{K+1}\|\mathbf{H}^{\mathrm{T}}-\mathbf{V}^{K+1}\mathbf{U}\|_F^2$$
$$s.t.\ \mathbf{U}_{i,j}\in\{0,1\},\ \sum_{i=1}^{C}\mathbf{U}_{i,j}=1 \quad (22)$$

Denoting $\mathbf{S}^{K+1}=\mathbf{H}$, the optimal solution of the cluster index matrix $\mathbf{U}$ at index $(i,j)$ can be easily obtained by the following formula [13]:

$$\mathbf{U}_{i,j}=\begin{cases}1, & i=\underset{c}{\operatorname{argmin}}\sum_{k}^{K+1}\alpha^k\left\|\mathbf{S}_{j,:}^{k,\mathrm{T}}-\mathbf{V}_{:,c}^k\right\| \\ 0, & otherwise\end{cases} \quad (24)$$

where $\mathbf{V}_{:,c}^k$ represents the clustering center of the $c$-th cluster in the $k$-th view.

## D. The Algorithm and Complexity Analysis

Based on the above analysis and derivations, the details of the proposed OMC-DR algorithm are given in Algorithm 1. Here, we provide the time complexity analysis of the algorithm as follows. Since the dimension of the common and specific representations, i.e., $m_c$ and $m_s$, are usually far less than the number of samples $N$ and the dimension of the original features $d^k$, the time complexity of both step 1 and step 2 are $O(N^2DT)$ and $O(N^2KDT)$, respectively, where $D$ is the maximal dimension of all the views, $K$ is the number of views and $T$ is the maximum number of iterations. The time complexity of both step 3 and 4 is $O\big((1+D)NKT\big)$. The time complexity of step 5 is $O((1+C)N^2T(1+K))$, where $C$ is the number of clusters. The time complexity of step 6 and step 7 is $O\big((1+CN+N)CT(1+K)\big)$ and $O(DNT)$, respectively. Since $D \gg C$, the overall time complexity of the algorithm is $O(N^2DKT)$.

Algorithms 1: OMC-DR

**Input:** multi-view data $\{\mathbf{X}^k\}$, $k$=1, 2, …, $K$, the dimension of specified and common representations $m_s$ and $m_c$, maximum number of iterations $T$, balance parameters $\beta,\ \gamma,\ \delta$.
**Initialization:** initialize $\{\mathbf{S}^k\}$, $\{\mathbf{W}^k\}$, $\{\mathbf{P}^k\}$, $\{\mathbf{V}^k\}$, $\mathbf{U}$ randomly and initialize $\alpha^k = 1/(K+1)$.
**for** $t$ form 1 to $T$ **do**
  1. update $\mathbf{H}$ by (11).
  **for** $k$ form 1 to $K$
    2. update $\mathbf{S}^k$ by (13).
    3: update $\mathbf{W}^k$ by (15).
    4: update $\mathbf{P}^k$ by (17).
  **end**
  **for** $k$ form 1 to $K$+1
    5. update $\alpha^k$ by (19).
    6. update $\mathbf{V}^k$ by (21).
  **end**
  7: update $\mathbf{U}$ by (23).
  8: Compute $Obj(t)$ by (9).
  **if** $t > 1$ **do**
    **if** $|Obj(t) - Obj(t-1)| < 1e-6$ **do**
      return.
    **end**
  **end**
**end**
**Output:** cluster indicator matrix $\mathbf{U}$.

## IV. Experimental Studies

In this section, the effeteness of the proposed method is evaluated. Specifically, Section IV-A gives the detailed experimental settings, Section IV-B gives the experimental results and analyses on real-world datasets, and Section IV-C further analyses and verifies the effectiveness of the specific representation and the unified clustering frame-work. The convergence and the runtime analyses of the proposed method is given in the Section IV-D. Finally, the statistical significance of the performance improvement achieved with the proposed method is examined in Section IV-E. The code of the proposed OMC-DR is available at https://github.com/BBKing49/OMC-DR.

### A. *Experimental Settings*

*1) Datasets*

Seven baseline multi-view datasets were adopted in our experiments, which are briefly described below. Table I gives the statistics of the datasets.

1) Reuters: It is a collection of documents translated into five languages, each of which is considered as a view [34]. English and French were selected as the two views of the experiments.

2) Cora: It is a publication dataset. Each publication is an example and consists of two views: a word vector of the main content and a word vector of the citation links pointing to this publication [34].

3) WebKB: It is a collection of the websites of four universities [25]. The website of one university was selected for the experiments. Page text, page hyperlink and title text were extracted as the three views.

4) 3Sources: It is a collection of stories gathered from three news websites [35]. Each story in three different websites as different views and 169 stories from all the three sources were selected for experiments.

5) Corel: It is an image classification dataset [35] with 10 classes. The SIFT and LBP features were extracted as the two views for the experiments.

6) Caltech7: It is an image dataset [6] and 1474 images were selected for the experiments. The GIST, CENTRIST, HOG and LBP features were extracted as four views.

7) NUS-WIDE: It is an image classification dataset [27] and 7 classes of images were selected for the experiments, in which the CORR and bag of visual words were extracted as two views.

Table I Statistics of the datasets

| **Datasets** | **Size** | **Views/Dimension** | **Class** |
|---|---|---|---|
| Reuters | 1200 | 2(2000-2000) | 6 |
| Cora | 2708 | 2(2708-1433) | 7 |
| WebKB | 226 | 3(2500-215-389) | 4 |
| 3Sources | 169 | 3(3560-3631-3068) | 6 |
| Corel | 1000 | 2(256-300) | 10 |
| Caltech7 | 1474 | 4(254-512-928-1984) | 7 |
| NUS-WIDE | 1000 | 2(144-500) | 5 |

Table II Parameter settings for nine algorithms

| Algorithms | Parameters and grid search ranges |
|---|---|
| TW-k-Means | Regularization parameter $\lambda$: {1,2, ..., 30}. Regularization parameter $\eta$: {10, 20, 30, ..., 120}. |
| GMultiNMF | Regularization parameters $\lambda_1$, $\lambda_2$: {$10^{-3}$, $10^{-2}$, …, $10^{3}$}. Dimension of common representation: the number of clusters. |
| NMFCC | Regularization parameters α, $\beta$, $\gamma$, $\mu$: {$10^{-3}$, $10^{-2}$, …, $10^{3}$}. Dimension of common representation: the number of clusters. |
| LMSC | Regularization parameters $\lambda$: {$10^{-3}$, $10^{-2}$, …, $10^{3}$}. Dimension of common representation: {10, 20, …, 90, 100}. |
| PLCMF | Regularization parameters $\beta$, $\gamma$, $\delta$, $\lambda$: {$10^{-3}$, $10^{-2}$, …, $10^{3}$}. Dimension of common representation: the number of clusters. |
| COMVSC | Regularization parameter $\lambda$ :{$2^3$, $2^5$, …, $2^{11}$, $2^{13}$}, Regularization parameter $\gamma$: {1.3, 1.5, …,2.5, 2.7} |
| CSMSC | Regularization parameters $\lambda_1$, $\lambda_2$: {$10^{-3}$, $10^{-2}$, …, $10^{3}$}. |
| MV-Co-VH | Regularization parameter $\eta$ : {$10^{-3}$, $10^{-2}$, …, $10^{3}$}. Collaborative learning parameter: {0.2, 0.4, …, 0.8, 1}. |
| OMC-DR (ours) | Regularization parameters $\beta$, $\gamma$, $\delta$: {$10^{-3}$, $10^{-2}$, …, $10^{3}$}. Dimension of common representation: the number of clusters. Dimension of the specific representation: {10, 20, …, 90, 100}. |

*2) Methods for Comparison*

Eight representative multi-view clustering algorithms were compared with the proposed OMC-DR. The algorithms are briefly described as follows:

1) TW-k-Means [1]: It is a multi-view clustering method based on the original views and extends the traditional K-means algorithms to the corresponding multi-view versions.

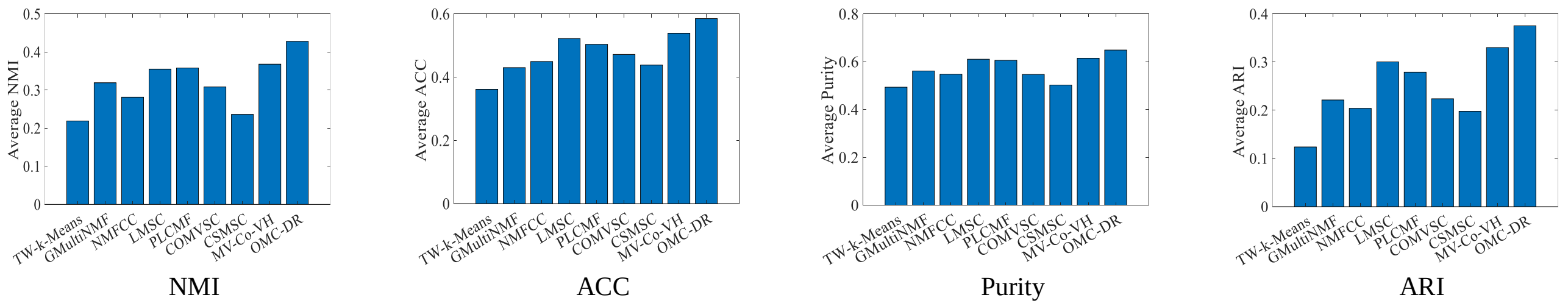

Fig. 3 The average results of nine algorithms on all datasets.

2) GMultiNMF [20]: It is a multi-view clustering method based on the common view, where NMF and manifold regularization are used to learn the common view for clustering.

3) NMFCC [28]: It is a multi-view clustering method based on the common view, where NMF and orthogonality regularization are used to learn the common view for clustering.

4) LMSC [8]: It is a multi-view clustering method based on the common view, where self-representation learning is used to learn the common view for clustering.

5) PLCMF [17]: It is a one-step multi-view clustering method, using matrix factorization to integrate common view learning with clustering partition as one single process that is guided by the pseudo label.

6) COMVSC [18]: It is a one-step multi-view clustering method, using self-representation learning to integrate subspace learning, partition fusion and spectral rotation into one process.

7) CSMSC [14]: This method uses the self-representation learning to explore the consistent and unique knowledge simultaneously, and then combines them to implement spectral clustering.

8) MV-Co-VH [16]: This method learns a common view and integrates it with other original views to realize collaborative learning for multi-view clustering.

*3) Parameters setting*

The hyperparameters of the nine algorithms of the experiments and the corresponding grid search ranges of the parameters involved are shown in Table II.

*4) Evaluation Indices*

Four commonly used evaluation indices, i.e., Normalized Mutual Information (NMI), Accuracy (ACC), Purity and Adjust Rand Indicator (ARI) were adopted in our experiments [15], [36]. Meanwhile, each algorithm was executed 30 times using different parameters with the grid search ranges specified. The best results in terms of the mean and standard deviation were recorded for comparison.

### *B. Experimental Results and Analysis*

The results, in terms of the four evaluation indices, obtained by the nine methods on all datasets are reported in Table III. The average performance is also visualized with the bar charts in Fig. 3. The following observations can be obtained from these results.

(1) For most datasets, the proposed OMC-DR has shown the best performance. This indicates the effectiveness of the proposed one-step learning mechanism and dual representation learning mechanism.

(2) For TW-k-Means which is based on the original views, its performance is much poorer than several common latent view-based methods, such as GMultiNMF, NMFCC and LMSC. This indicates that the original visual views are not sufficient to obtain ideal clustering results for multi-view data.

(3) For the three methods based on the common latent view, i.e., GMultiNMF, NMFCC and LMSC, their performance is better than that of TW-k-Means method which is based on the original views. It indicates that common latent information of multiple views is advantageous over the original views for multi-view clustering.

(4) For methods based on the common latent view, by comparing the three two-step methods (i.e., GMultiNMF, NMFCC and LMSC) with the three one-step methods (PLCMF, COMVSC and OMC-DR), we can see that the performance of the latter is better. It indicates that separating representation learning and clustering partition is inferior strategy to the integration of them as a single process.

(5) Comparing the three one-step multi-view clustering methods, i.e., OMC-DR, PLCMF and COMVSC, the performance of the proposed OMC-DR is the best among them for most of the datasets. This is because that PLCMF and COMVSC only explore the common latent information, while OMC-DR exploits both the common information of all views and the specific information of each view. This also indicates that both types of information are useful for clustering partition.

(6) For the two methods that use the common and specific representations, i.e., the proposed OMC-DR and CSMSC, better results are obtained by the former, which is attributed to the proposed mechanism that can reduce the redundancy between the common and specific representations, and also due to the proposed one-step learning strategy that increases the relevance of the learning representation for clustering.

(7) For the MV-Co-VH method which is based on visible-hidden view cooperation, it has shown much better performance than the other methods. This is because the original visible views and the hidden common view have realized effective cooperation that make the common information and specific information more effective for clustering. However, the performance of MV-Co-VH is still inferior to the proposed OMC-DR, which is due to the stronger representation learning ability of OMC-DR as compared to MV-Co-VH.

Table III. Clustering results on seven datasets

| Datasets | Algorithms | NMI | ACC | Purity | ARI |
|---|---|---|---|---|---|
| Reuters | TW-k-Means | 0.2730±0.0206 | 0.4708±0.0309 | 0.4722±0.0296 | 0.1917±0.0143 |
| | GMultiNMF | 0.2122±0.0000 | 0.3917±0.0000 | 0.4100±0.0000 | 0.1522±0.0000 |
| | NMFCC | 0.2966±0.0134 | 0.4756±0.0294 | 0.4828±0.0234 | 0.1974±0.0143 |
| | LMSC | 0.3057±0.0020 | 0.4931±0.0029 | 0.5317±0.0106 | 0.2407±0.0042 |
| | PLCMF | 0.3322±0.0264 | 0.4892±0.0260 | 0.5307±0.0263 | 0.2304±0.0233 |
| | COMVSC | 0.1788±0.0017 | 0.3314±0.0009 | 0.3366±0.0004 | 0.0700±0.0004 |
| | CSMSC | 0.1501±0.0221 | 0.3756±0.0330 | 0.3764±0.0330 | 0.1170±0.0072 |
| | MV-Co-VH | 0.3006±0.0000 | 0.4610±0.0000 | 0.4988±0.0000 | 0.1897±0.0000 |
| | OMC-DR(Ours) | **0.3942±0.0003** | **0.5750±0.0018** | **0.5850±0.0002** | **0.3265±0.0006** |
| Cora | TW-k-Means | 0.1750±0.0498 | 0.2051±0.0548 | 0.3804±0.0459 | 0.0309±0.0302 |
| | GMultiNMF | 0.2084±0.0000 | 0.3508±0.0000 | 0.4298±0.0000 | 0.1183±0.0000 |
| | NMFCC | 0.1600±0.0117 | 0.3542±0.0178 | 0.4031±0.0181 | 0.0918±0.0141 |
| | LMSC | 0.2495±0.0060 | 0.4824±0.0058 | 0.5373±0.0076 | 0.2001±0.0058 |
| | PLCMF | 0.2760±0.0310 | 0.4729±0.0381 | 0.5123±0.0363 | 0.1238±0.0292 |
| | COMVSC | 0.0763±0.0066 | 0.3040±0.0181 | 0.3904±0.0136 | 0.0571±0.0098 |
| | CSMSC | 0.1659±0.0000 | 0.3040±0.0012 | 0.4078±0.0000 | 0.0762±0.0000 |
| | MV-Co-VH | **0.3153±0.0047** | **0.4971±0.0270** | **0.5520±0.0047** | **0.2476±0.0094** |
| | OMC-DR(Ours) | 0.2773±0.0001 | 0.4889±0.0020 | 0.5281±0.0008 | 0.2274±0.0031 |
| WebKB | TW-k-Means | 0.2084±0.0235 | 0.5310±0.0160 | 0.6372±0.0260 | 0.1046±0.0688 |
| | GMultiNMF | 0.4459±0.0000 | 0.5531±0.0000 | 0.7876±0.0000 | 0.2889±0.0000 |
| | NMFCC | 0.4683±0.0432 | 0.6091±0.0270 | 0.7994±0.0170 | 0.3389±0.0387 |
| | LMSC | 0.5004±0.0319 | 0.7950±0.0109 | 0.7950±0.0150 | 0.5955±0.0297 |
| | PLCMF | 0.4848±0.0732 | 0.7372±0.0109 | 0.8053±0.0000 | 0.5472±0.0109 |
| | COMVSC | 0.5678±0.0108 | 0.8024±0.0461 | 0.8024±0.0461 | 0.5597±0.0934 |
| | CSMSC | 0.3200±0.0018 | 0.6386±0.0102 | 0.6386±0.0100 | 0.2975±0.0273 |
| | MV-Co-VH | 0.4507±0.0907 | 0.7496±0.0943 | 0.7903±0.0590 | 0.5606±0.0312 |
| | OMC-DR(Ours) | **0.6476±0.0007** | **0.8673±0.0001** | **0.8673±0.0016** | **0.6711±0.0003** |
| 3Sources | TW-k-Means | 0.1634±0.0846 | 0.1420±0.0698 | 0.4142±0.0571 | 0.0183±0.0040 |
| | GMultiNMF | 0.4123±0.0000 | 0.4142±0.0000 | 0.6213±0.0000 | 0.2386±0.0000 |
| | NMFCC | 0.2586±0.1345 | 0.4339±0.0964 | 0.5365±0.0917 | 0.1416±0.1610 |
| | LMSC | 0.5655±0.0150 | 0.5424±0.0068 | 0.7396±0.0059 | 0.4004±0.0203 |
| | PLCMF | 0.4685±0.0788 | 0.5065±0.0136 | 0.6769±0.0553 | 0.2992±0.0537 |
| | COMVSC | 0.5074±0.0000 | 0.5621±0.0002 | 0.7101±0.0032 | 0.3624±0.0036 |
| | CSMSC | 0.3443±0.0085 | 0.5779±0.0090 | 0.6114±0.0090 | 0.3021±0.0155 |
| | MV-Co-VH | 0.5843±0.0248 | 0.7243±0.0256 | 0.7716±0.0208 | 0.5559±0.0387 |
| | OMC-DR(Ours) | **0.6927±0.0010** | **0.7456±0.0014** | **0.8255±0.0000** | **0.6164±0.0009** |
| Corel | TW-k-Means | 0.2236±0.0183 | 0.3310±0.0090 | 0.3443±0.0221 | 0.1325±0.0026 |
| | GMultiNMF | 0.2661±0.0000 | 0.3170±0.0000 | 0.3690±0.0000 | 0.1441±0.0000 |
| | NMFCC | 0.2706±0.0016 | 0.3460±0.0079 | 0.3940±0.0020 | 0.1531±0.0038 |
| | LMSC | 0.2892±0.0021 | 0.3870±0.0155 | 0.4253±0.0110 | 0.1776±0.0044 |
| | PLCMF | 0.2881±0.0180 | 0.3782±0.0281 | 0.4184±0.0274 | 0.1848±0.0260 |
| | COMVSC | 0.2452±0.0043 | 0.3180±0.0199 | 0.3377±0.0125 | 0.1311±0.0199 |
| | CSMSC | 0.1990±0.0030 | 0.3077±0.0095 | 0.3363±0.0087 | 0.1073±0.0041 |
| | MV-Co-VH | 0.2988±0.0066 | 0.3856±0.0288 | 0.4188±0.0174 | 0.1934±0.0196 |
| | OMC-DR(Ours) | **0.3202±0.0022** | **0.3970±0.0007** | **0.4370±0.0000** | **0.1967±0.0005** |
| Caltech7 | TW-k-Means | 0.3734±0.0027 | 0.4532±0.0324 | 0.7944±0.0129 | 0.2916±0.0053 |
| | GMultiNMF | 0.5591±0.0000 | 0.5909±0.0000 | 0.9064±0.0000 | 0.5121±0.0000 |
| | NMFCC | 0.4140±0.0405 | 0.5330±0.0614 | 0.8342±0.0024 | 0.4219±0.1115 |
| | LMSC | 0.4663±0.0085 | 0.5663±0.0146 | 0.8503±0.0051 | 0.3976±0.0051 |
| | PLCMF | **0.5475±0.0585** | **0.5917±0.0114** | **0.8950±0.0221** | **0.4811±0.0119** |
| | COMVSC | 0.4427±0.0028 | 0.5588±0.0012 | 0.8280±0.0014 | 0.2694±0.0036 |
| | CSMSC | 0.4379±0.0037 | 0.5749±0.0079 | 0.8535±0.0041 | 0.4624±0.0149 |
| | MV-Co-VH | 0.5136±0.0510 | 0.5627±0.0104 | 0.8708±0.0122 | 0.4743±0.0115 |
| | OMC-DR(Ours) | 0.5209±0.0004 | 0.5834±0.0002 | 0.8623±0.0031 | 0.4725±0.0007 |
| NUS-WIDE | TW-k-Means | 0.1157±0.0089 | 0.3973±0.0259 | 0.4133±0.0055 | 0.0980±0.0095 |
| | GMultiNMF | 0.1311±0.0000 | 0.3930±0.0000 | 0.4070±0.0000 | 0.0962±0.0000 |
| | NMFCC | 0.1020±0.0070 | 0.3743±0.0081 | 0.3857±0.0110 | 0.0837±0.0038 |
| | LMSC | 0.1088±0.0016 | 0.3893±0.0029 | 0.3957±0.0023 | 0.0918±0.0032 |
| | PLCMF | 0.1092±0.0082 | 0.3498±0.0076 | 0.4044±0.0183 | 0.0877±0.0101 |
| | COMVSC | 0.1402±0.0104 | 0.4253±0.0021 | 0.4333±0.0090 | **0.1183±0.0059** |
| | CSMSC | 0.0356±0.0000 | 0.2892±0.0053 | 0.2940±0.0010 | 0.0234±0.0000 |
| | MV-Co-VH | 0.1131±0.0047 | 0.3920±0.0148 | 0.4028±0.0101 | 0.0901±0.0056 |
| | OMC-DR(Ours) | **0.1425±0.0020** | **0.4390±0.0011** | **0.4390±0.0006** | 0.1181±0.0003 |

Table IV. Performance of OMC-DR, OMC-DR _NS and OMC-DR_NC on all datasets

| Datasets | OMC-DR _NC | | | | OMC-DR _NS | | | | OMC-DR | | | |
|---|---|---|---|---|---|---|---|---|---|---|---|---|
| | NMI | ACC | Purity | ARI | NMI | ACC | Purity | ARI | NMI | ACC | Purity | ARI |
| Reuters | 0.2835 ±0.0037 | 0.4817 ±0.0037 | 0.5108 ±0.0019 | 0.2368 ±0.0110 | **0.4083 ±0.0061** | **0.5983 ±0.0087** | **0.5983 ±0.0037** | 0.3145 ±0.0006 | 0.3942 ±0.0003 | **0.5750 ±0.0018** | **0.5850 ±0.0002** | **0.3265 ±0.0006** |
| Cora | 0.1349 ±0.0005 | 0.3397 ±0.0130 | 0.4014 ±0.0033 | 0.0925 ±0.0015 | 0.2663 ±0.0012 | **0.4904** ±0.0017 | 0.5148 ±0.0061 | 0.2186 ±0.0041 | **0.2773 ±0.0001** | 0.4889 ±0.0020 | 0.5281 ±0.0008 | **0.2274 ±0.0031** |
| WebKB | 0.4807 ±0.0008 | 0.5752 ±0.0063 | 0.8186 ±0.0032 | 0.4184 ±0.0102 | 0.5810 ±0.0036 | 0.6195 ±0.0035 | 0.8451 ±0.0013 | 0.6645 ±0.0028 | **0.6476 ±0.0007** | **0.8673 ±0.0001** | **0.8673 ±0.0016** | **0.6711 ±0.0003** |
| 3Sources | 0.5876 ±0.0050 | 0.6509 ±0.0042 | 0.7633 ±0.0031 | 0.5215 ±0.0030 | 0.6264 ±0.0061 | 0.6970 ±0.0037 | 0.7041 ±0.0023 | 0.5225 ±0.0043 | **0.6927 ±0.0010** | **0.7456 ±0.0014** | **0.8255 ±0.0000** | **0.6164 ±0.0009** |
| Corel | 0.2693 ±0.0101 | 0.3090 ±0.0061 | 0.3530 ±0.0011 | 0.1347 ±0.0042 | 0.2845 ±0.0021 | 0.3230 ±0.0073 | 0.3940 ±0.0008 | 0.1529 ±0.0033 | **0.3202 ±0.0022** | **0.3970 ±0.0007** | **0.4370 ±0.0000** | **0.1967 ±0.0005** |
| Caltech7 | 0.4429 ±0.0031 | 0.4647 ±0.0010 | 0.8433 ±0.0035 | 0.3268 ±0.0041 | 0.5136 ±0.0031 | 0.5556 ±0.0078 | 0.8569 ±0.0093 | 0.4656 ±0.0008 | **0.5209 ±0.0004** | **0.5834 ±0.0002** | **0.8623 ±0.0031** | **0.4725 ±0.0007** |
| NUS-WIDE | 0.1131 ±0.0052 | 0.4010 ±0.0018 | 0.4010 ±0.0011 | 0.0860 ±0.0014 | 0.1329 ±0.0120 | 0.3930 ±0.0017 | 0.4010 ±0.0037 | 0.1090 ±0.0010 | **0.1425 ±0.0020** | **0.4390 ±0.0011** | **0.4390 ±0.0006** | **0.1181 ±0.0003** |

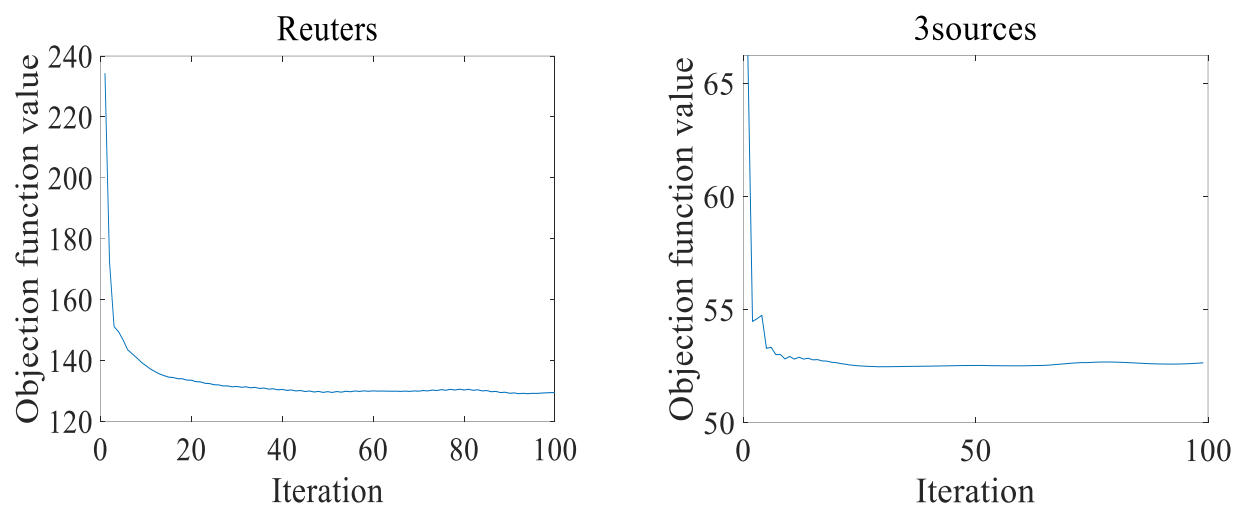

(a) (c)

Fig. 4: Convergence of OMC-DR on the Reuter and 3source datasets.

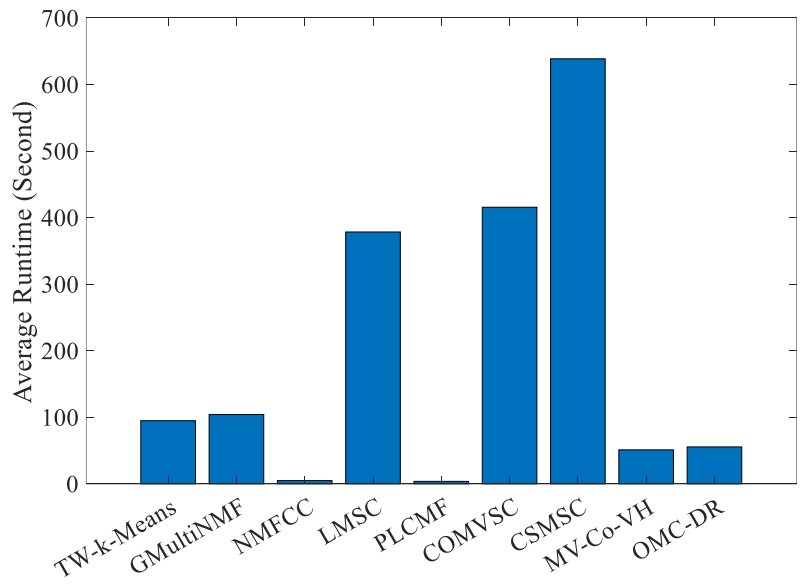

Fig. 5 Average runtime of the nine algorithms on all datasets

Table V. The *p*-value and ranking of the nine algorithms based on NMI

| Algorithms | Ranking | *p* -value | Null Hypothesis |
|---|---|---|---|
| OMC-DR | 1.2857 | 0.000034 | Rejected |
| MV-Co-VH | 3.1429 | | |
| PLCMF | 3.5714 | | |
| LMSC | 4 | | |
| COMVSC | 5.2857 | | |
| GMultiNMF | 6.1429 | | |
| NMFCC | 6.5714 | | |
| TW-k-Means | 7.1429 | | |
| CSMSC | 7.8571 | | |

Table VI. The *p*-value and ranking of the nine algorithms based on ACC

| Algorithms | Ranking | *p* -value | Null Hypothesis |
|---|---|---|---|
| OMC-DR | 1.5 | 0.002121 | Rejected |
| LMSC | 3.8571 | | |
| MV-Co-VH | 3.8571 | | |
| PLCMF | 4.5714 | | |
| COMVSC | 4.7857 | | |
| GMultiNMF | 6.2857 | | |
| NMFCC | 6.2857 | | |
| CSMSC | 6.7143 | | |
| TW-k-Means | 7.1429 | | |

### *C. Effectiveness Analysis*

We further compare OMC-DR with two special versions of the algorithm to evaluate the effectiveness of the proposed dual representation learning mechanism and one-step learning mechanism. The two special versions of OMC-DR, denoted by OMC-DR_NS and OMC-DR_NC, refer to OMC-DR without considering specific representation and OMC-DR that separates representation learning from clustering partition, respectively. Specifically, the objective functions of OMC-DR_NS and OMC-DR_NC are defined as follows:

$$\min_{\mathbf{H},\mathbf{W}^k,\mathbf{V},\mathbf{U}} \sum_{k=1}^{K}\|\mathbf{X}^k-\mathbf{H}\mathbf{W}^k\|_F^2+\gamma\|\mathbf{H}\|_{2,1}^2+\|\mathbf{H}^{\mathrm{T}}-\mathbf{V}\mathbf{U}\|_F^2$$
$$s.t.\ \mathbf{U}_{i,j}\in\{0,1\},\ \sum_{i=1}^{C}\mathbf{U}_{i,j}=1 \quad (24)$$

$$\min_{\mathbf{H},\mathbf{W}^k,\mathbf{S}^k,\mathbf{P}^k} \sum_{k=1}^{K}\|\mathbf{X}^k-\mathbf{H}\mathbf{W}^k-\mathbf{S}^k\mathbf{P}^k\|_F^2+\gamma\|\mathbf{H}\|_{2,1}^2+$$
$$\gamma\|\mathbf{S}^k\|_{2,1}^2+\beta\sum_{k=1}^{K}\|\mathbf{H}^{\mathrm{T}}\mathbf{S}^k\|_F^2 \quad (25)$$

An alternating optimization scheme is used to solve the objective functions in (24) and (25). Besides, the learned common and specific representations in OMC-DR_NC are combined as input of the K-Means algorithm to get the final clustering results. The experimental results of the three versions on all the datasets are presented in Table IV. It can be seen that the performance of OMC-DR_NC is inferior to OMC-DR and OMC-DR_NS, indicating that one-step mechanism is very useful to improve the clustering performance. In addition, although there is little difference between OMC-DR_NS and OMC-DR on the Reuters and Caltech7 datasets, OMC-DR is advantageous to OMC-DR_NS on the other datasets, which shows that both the common information of different views and the specific information of each view are useful for improving the clustering effect.

### *D. Runtime and Convergence Analysis*

In this subsection, the convergence and the computational cost of the proposed method are analyzed. First, we conduct the convergence analysis based on the Reuters and 3Sources datasets. Fig. 4 plots the value of the objective function at different iteration steps. From the results, we find that the value decreases rapidly with the iterations on the two datasets, indicating that convergence can usually be reached within 50 iterations.

Table VII. Post-hoc Holm results based on NMI (reject hypothesis if $p$-value <0.016667)

| $i$ | Algorithm | $z=(R_o-R_i)/SE$ | $p$-value | Holm $=\alpha/i, \alpha=0.05$ | Null Hypothesis |
|---|---|---|---|---|---|
| 8 | CSMSC /OMC-DR | 4.48914 | 0.000007 | 0.00625 | Rejected |
| 7 | TW-k-Means /OMC-DR | 4.00119 | 0.000063 | 0.007143 | Rejected |
| 6 | NMFCC /OMC-DR | 3.61083 | 0.000305 | 0.008333 | Rejected |
| 5 | GMultiNMF /OMC-DR | 3.31806 | 0.000906 | 0.01 | Rejected |
| 4 | COMVSC /OMC-DR | 2.73252 | 0.006285 | 0.0125 | Rejected |
| 3 | LMSC /OMC-DR | 1.85421 | 0.063709 | 0.016667 | Not Rejected |
| 2 | PLCMF /OMC-DR | 1.56144 | 0.11842 | 0.025 | Not Rejected |
| 1 | MV-Co-VH /OMC-DR | 1.26867 | 0.204559 | 0.05 | Not Rejected |

Table VIII. Post-hoc Holm results based on ACC (reject hypothesis if $p$-value <0.0125)

| $i$ | Algorithm | $z=(R_o-R_i)/SE$ | $p$-value | Holm $=\alpha/i, \alpha=0.05$ | Null Hypothesis |
|---|---|---|---|---|---|
| 8 | TW-k-Means /OMC-DR | 3.854805 | 0.000116 | 0.00625 | Rejected |
| 7 | CSMSC /OMC-DR | 3.562035 | 0.000368 | 0.007143 | Rejected |
| 6 | GMultiNMF /OMC-DR | 3.169265 | 0.000978 | 0.008333 | Rejected |
| 5 | NMFCC /OMC-DR | 3.169265 | 0.000978 | 0.01 | Rejected |
| 4 | COMVSC /OMC-DR | 2.24457 | 0.014796 | 0.0125 | Not Rejected |
| 3 | PLCMF /OMC-DR | 2.09185 | 0.035889 | 0.016667 | Not Rejected |
| 2 | LMSC /OMC-DR | 1.610235 | 0.107347 | 0.025 | Not Rejected |
| 1 | MV-Co-VH /OMC-DR | 1.610235 | 0.107347 | 0.05 | Not Rejected |

Second, we compare the average runtime of the nine algorithms on all datasets. As shown in Fig. 5, the runtime of OMC-DR is at the middle to upper level, which is faster than five out of the eight methods under comparison. Although OMC-DR has no advantage over NMFCC, PLCMF and MV-Co-VH in terms of runtime, it has obvious advantages over them in clustering performance in terms of the evaluation indices as shown in Table III and Fig. 3.

### E. *Statistical Analyses*

In this subsection, statistical analysis is conducted to further investigate the performance of the nine algorithms by running the Friedman test and the post-hoc Holm test.

First, we evaluate whether the clustering results of the nine algorithms exhibit statistically significant differences using the Friedman test [37]. The null hypothesis of the Friedman test is that there is not difference in clustering results between all methods. When the $p$-value of the test is less than 0.05, the null hypothesis is rejected, i.e., the difference in clustering results is significant. Tables V and VI show the Friedman test results of the nine methods based on NMI and ACC, respectively. For the Friedman test results based on Purity and ARI are shown in Tables S1-S2 in the *Supplementary Materials* section, respectively. As shown in the tables, the difference in performance between these methods is significant. Moreover, OMC-DR is ranked first (smallest ranking values), which means that the proposed OMC-DR is the best among all methods.

Next, to verify whether the performance of OMC-DR is significantly better than each of the other eight methods, we conduct the post-hoc Holm test. Tables VII and VIII show the results based on NMI and ACC respectively, and the results based on Purity and ARI are shown in Tables S3-S4 in the *Supplementary Materials* section. Table VII shows that the performance difference between OMC-DR and the following five methods, CSMSC, TW-k-Means, NMFCC, GMultiNMF and COMVSC, is significant. Table VIII shows that the performance difference between OMC-DR and the following four methods, TW-k-Means, CSMSC, GMultiNMF and NMFCC, is significant, whereas Table S3-S4 shows respectively that the difference between OMC-DR and four, five algorithms is statistically significant. Although the post-hoc Holm test shows that the performance advantage of the proposed OMC-DR over some methods, such as MV-Co-VH, is not significant, the results in Table III and Fig. 3 reveal that OMC-DR still outperforms them to a certain extent on various evaluation indices.

## V. CONCLUSIONS

In this paper, we propose a novel multi-view clustering method by exploring the dual representation of the common and the specific information of different views simultaneously. Meanwhile, we design a one-step framework to integrate dual representation learning and the cluster indicator matrix learning into a single process, by which the two learning tasks can mutually benefit from each other to achieve a better clustering result. An alternating optimization scheme is developed to solve the optimization problem of the propose method. Experimental results on real world multi-view datasets have demonstrated the superiority of the proposed method.

Although the proposed method has achieved outstanding performance, there are still rooms for the further improvement. For example, the proposed method only explores the linear relationship of the data. It could be more effective if the nonlinear relationship is exploited, by introducing such mechanisms as deep learning technique and kernel method. Besides, the proposed method is not applicable for incomplete multi-view data, which is a common issue in real applications. This will be tackled in detail in our future work.